\documentclass[conference]{IEEEtran}
\usepackage[letterpaper,left=0.595in,right=0.595in,top=0.75in,bottom=1.013in]{geometry}
\usepackage{amsmath}
\usepackage{graphicx}
\usepackage{amssymb}
\usepackage{cite}
\usepackage{makecell}
\usepackage{multirow}
\usepackage[capitalise]{cleveref} 
\usepackage{textgreek} 
\usepackage{newtxtext,newtxmath}
\usepackage[utf8]{inputenc}
\usepackage[capitalise]{cleveref} 
\usepackage{footnote}

\IEEEoverridecommandlockouts
\IEEEaftertitletext{\vspace{-1cm}}

\title{ Multi-Channel Swin Transformer Framework for Bearing Remaining Useful Life Prediction}

\author{
\hspace{-2cm}
\IEEEauthorblockN{Ali Mohajerzarrinkelk}
    \IEEEauthorblockA{
          \hspace{-2cm} Faculty of Engineering \\
          \hspace{-1.9cm} and Computer Science \\
          \hspace{-1.9cm} University of Victoria, Canada \\
          \hspace{-1.9cm} alimohajer@uvic.ca
    }
\and
\IEEEauthorblockN{Maryam Ahang}
\IEEEauthorblockA{
         Faculty of Engineering \\
          and Computer Science \\
         University of Victoria, Canada \\
         maryamahang@uvic.ca
    }
\and
\IEEEauthorblockN{Mehran Zoravar}
\IEEEauthorblockA{
         Faculty of Engineering \\
          and Computer Science \\
         University of Victoria, Canada \\
         mehranzoravar@uvic.ca
    }
\and
\hspace{9cm}
\IEEEauthorblockN{
    \begin{tabular}{c} 
        \hspace{-6.25cm} Mostafa Abbasi
    \end{tabular}}
\IEEEauthorblockA{
        Faculty of Engineering \\
         and Computer Science \\
        University of Victoria, Canada \\
    abbasi@uvic.ca
}
\and
\IEEEauthorblockN{Homayoun Najjaran}
\IEEEauthorblockA{
        Faculty of Engineering \\
         and Computer Science \\
        University of Victoria, Canada \\
        najjaran@uvic.ca
    }

}

\begin{document}
\maketitle

\begin{abstract}
Precise estimation of the Remaining Useful Life (RUL) of rolling bearings is an important consideration to avoid unexpected failures, reduce downtime, and promote safety and efficiency in industrial systems. Complications in degradation trends, noise presence, and the necessity to detect faults in advance make estimation of RUL a challenging task. This paper introduces a novel framework that combines wavelet-based denoising method, Wavelet Packet Decomposition (WPD), and a customized multi-channel Swin Transformer model (MCSFormer) to address these problems.  With attention mechanisms incorporated for feature fusion, the model is designed to learn global and local degradation patterns utilizing hierarchical representations for enhancing predictive performance. Additionally, a customized loss function is developed as a key distinction of this work to differentiate between early and late predictions, prioritizing accurate early detection and minimizing the high operation risks of late predictions. The proposed model was evaluated with the PRONOSTIA dataset using three experiments. Intra-condition experiments demonstrated that MCSFormer outperformed state-of-the-art models, including the Adaptive Transformer, MDAN, and CNN-SRU, achieving 41\%, 64\%, and 69\% lower MAE on average across different operating conditions, respectively. In terms of cross-condition testing, it achieved superior generalization under varying operating conditions compared to the adapted ViT and Swin Transformer. Lastly, the custom loss function effectively reduced late predictions, as evidenced in a 6.3\% improvement in the scoring metric while maintaining competitive overall performance. The model's robust noise resistance, generalization capability, and focus on safety make MCSFormer a trustworthy and effective predictive maintenance tool in industrial applications.

\end{abstract}


\section{Introduction}
Accurately predicting the RUL of rolling bearings is critical for modern predictive maintenance frameworks. Bearings are vital components in rotary machines, and their reliability directly impacts operational safety, efficiency, and cost-effectiveness. Traditional methods for RUL prediction, such as physics-based models and statistical techniques \cite{cubillo2016}, provided foundational insights but struggled to generalize across varying operating conditions and complex fault development trajectories, particularly in dynamic industrial environments \cite{althubaiti2022}.

The advent of data-driven approaches addressed many of these challenges. While early machine learning techniques improved prediction accuracy in some scenarios, their ability to detect complex degradation trends was low, and they were more sensitive to noise, thus reducing their potential to be applied in real-world applications \cite{ahang2024}. Deep learning methods, particularly Convolutional Neural Networks (CNNs), transformed RUL prediction by learning features automatically from raw vibration signals \cite{deng2022}. CNNs, nonetheless, struggled with temporal dependencies, and thus Long Short-Term Memory (LSTM) networks were adopted \cite{song2024}. Transformer models \cite{vaswani2017attention} later advanced sequence modeling using self-attention to learn short- and long-term dependencies without sequential processing, making them extremely powerful for time-series data. Su et al. \cite{su2021} enhanced RUL estimation by integrating transformers with pre-extraction of features, which improved accuracy with diverse operating conditions.

However, robust feature extraction techniques are essential to enhance the model’s ability to isolate informative characteristics contributing to fault progression. Wavelet-based techniques have been used to improve feature extraction for predictive maintenance tasks \cite{ali2019induction}. In contrast to Fourier transforms, wavelet transforms allow localized analysis in the time-frequency domain and are particularly beneficial for detecting transient changes in vibration signals, which often indicate early-stage degradation. WPD enhances conventional wavelet transforms by allowing both low- and high-frequency band decomposition and hence providing a more detailed representation of intricate degradation patterns \cite{gao2024}.  However, the effectiveness of WPD is heavily reliant on the quality of the input signal. Noise can obscure critical time-frequency features, reducing the accuracy of predictive models. To address this, robust denoising techniques, such as wavelet-based denoising with adaptive soft thresholding \cite{donoho1995}, are crucial for preserving informative transient patterns while eliminating redundant or high-frequency noise. Providing a denoised signal into the model ensures WPD-derived features remain interpretable and robust, and final RUL predictions become more reliable.

Hybrid approaches improve RUL prediction through the combination of the strengths of different deep learning models. Transformer-CNN hybrid models improve the representation of complex degradation dynamics by combining convolutional feature extraction with attention-based sequence modeling  \cite{guo2022}. However, such hybrid architectures tend to depend on convolutional layers with fixed receptive fields, limiting adaptability to varying fault progression trends. To mitigate this, Vision Transformers (ViTs) \cite{dosovitskiy2020image} remove convolutional operations through patch-based self-attention, resulting in a more flexible and generalizable feature representation. Although effective, ViTs process images as non-overlapping patches, potentially losing fine-grained spatial information important for accurate RUL prediction. Swin Transformers \cite{liu2021swin} overcome this drawback by introducing a hierarchical structure that utilizes shifted window-based self-attention. This design enables effective multi-scale feature extraction, capturing both local and global dependencies while maintaining spatial continuity across image regions. As a result, Swin Transformers offer a more balanced approach, preserving important structural information while leveraging the strengths of transformer architectures. Although ViTs eliminate convolutional operations, structured preprocessing through convolutional layers continues to be advantageous in refining feature representations prior to feeding them into transformer-based models. 

In RUL prediction, timing is just as crucial as accuracy. Early predictions, where the predicted RUL is shorter than the actual RUL, can lead to unnecessary maintenance, increasing costs and disrupting operations. Late predictions, where RUL is longer than the actual RUL, pose serious safety risks by postponing essential maintenance, potentially causing unexpected failure or catastrophic damage. Despite their advancements, existing approaches often fail to weigh the risks of early and late predictions separately, highlighting the need for a framework that aims to minimize late predictions to ensure safety while avoiding unnecessary early interventions that may increase maintenance costs.

To address the challenges mentioned, a novel framework called MCSFormer is introduced in this paper. At the first stage, vibration signals undergo wavelet-based denoising process to preserve critical fault characteristics. The denoised signals are then projected onto time-frequency representations through WPD, preserving intricate frequency-domain features. After that, there is a convolutional processing phase that enhances the learned features prior to their input to a Swin Transformer backbone, capable of efficiently capturing spatial and temporal relationships. The hierarchical attention mechanism enhances feature interactions between horizontal and vertical inputs, while a multi-scale regression head refines the final prediction. 
To effectively address the issue of late predictions, a custom loss function is employed, introducing an additional penalty specifically targeting instances where the predicted RUL exceeds the actual RUL during training. With the combination of noise reduction, feature extraction, and hierarchical attention, MCSFormer is an end-to-end method for reliable RUL prediction. 
The rest of the paper is organized as follows: Section II elaborates on the proposed method with data preprocessing, model structure, and loss function design. Section III contains experimental tests and results, and Section IV concludes the study with insights and future research directions.

\section{Methodology}

\begin{figure*}[tb]
    \centering
    \resizebox{0.85\linewidth}{!}{\includegraphics{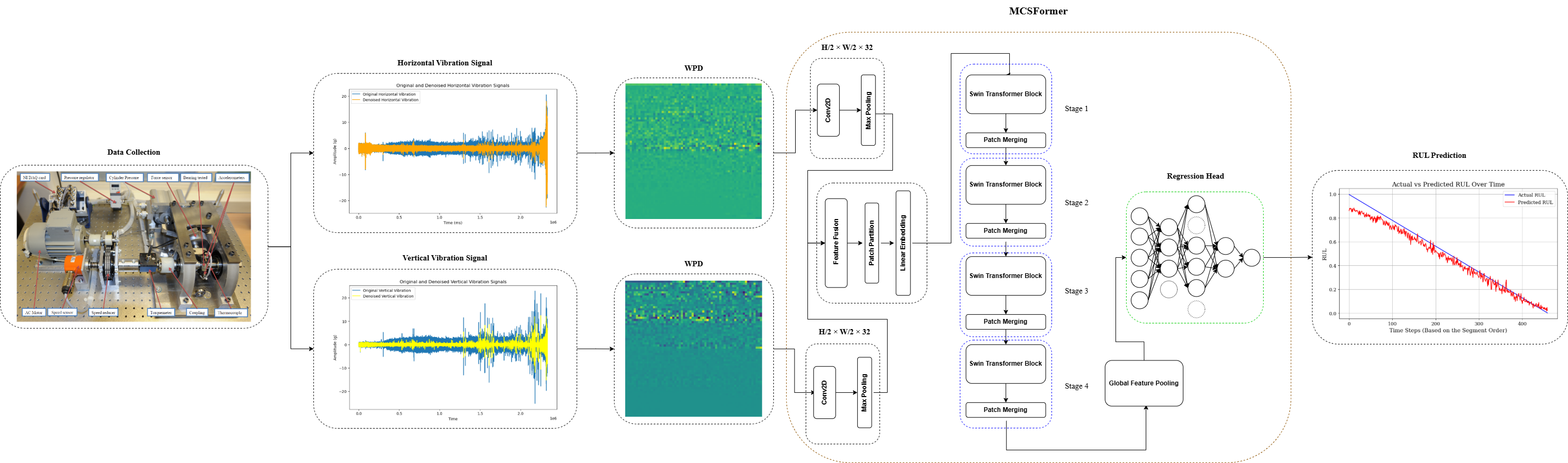}}
    \caption{Overview of the framework}
    \label{fig:methodology_framework}
\end{figure*}

In this work, we present MCSFormer, a robust framework designed to accurately predict the RUL of rolling bearings. The framework integrates wavelet-based denoising, sliding window processing, WPD, and a Multi-channel Swin Transformer architecture to address challenges in noise resilience, degradation modeling, and early fault detection. The overall methodology is illustrated in Figure~\ref{fig:methodology_framework}, which outlines the sequential steps of the proposed framework.

\subsection{Dataset}
The dataset used in this study originates from the PRONOSTIA experimental platform, which provides run-to-failure vibration data from 17 bearings tested under three distinct operating conditions. Each vibration signal is sampled at 25.6 kHz over 0.1-second intervals, with a 10-second gap between measurements \cite{nectoux2012pronostia}.

\subsection{Data Preprocessing}
The preprocessing pipeline is critical for enhancing signal quality and extracting meaningful features for modeling. For this framework, the preprocessing pipeline is as follows:

\subsubsection{Denoising Methods}
Denoising is performed in two sequential steps to isolate fault signatures and suppress noise:
\begin{enumerate}
    \item[{I.}] {Wavelet-Based Denoising:} The vibration signal is decomposed using the db5 wavelet into approximation and detail coefficients up to level 2. Adaptive soft-thresholding is applied to the detail coefficients, suppressing noise while retaining transient fault-related features \cite{donoho1995}.
    \item[{II.}] {Savitzky-Golay Filtering:} Polynomial smoothing is applied to the denoised signal using a window size of 5 and a polynomial order of 2, maintaining transient features while reducing abrupt fluctuations \cite{schafer2011}.
\end{enumerate}

\subsubsection{Sliding Window Segmentation}
The vibration signals collected at fixed intervals are segmented with a sliding window approach to effectively acquire temporal correlations. A total of ten sets of consecutive measurement intervals (window size) are contained in each window, thereby preserving key temporal patterns in the degradation process without incurring high computational costs. A stride of five intervals is applied, which means there is a 50\% overlap between successive windows.

\subsubsection{Wavelet Packet Decomposition (WPD)}
The denoised signals are subsequently applied with WPD to extract detailed time-frequency representations, which can effectively capture low- and high-frequency fault features. The commonly used db5 wavelet is employed as the basis function in this study. Signals are decomposed up to level 3, generating wavelet packet coefficients, which are then reshaped into 64×64 2D images as input to the Swin Transformer model.

\subsubsection{First Prediction Time (FPT) Determination and Data Labeling}

\paragraph{FPT Determination}
The FPT marks the transition from a healthy to a degraded state and is determined using kurtosis analysis. Kurtosis is calculated for the healthy stage of the bearing's life. The FPT is identified when \(K\) exceeds the threshold \(\mu \pm 3\sigma\) of the healthy stage's kurtosis for three consecutive data points \cite{gao2024, song2024}.

\paragraph{Labeling Strategy}
Data prior to the FPT is removed from the training data, as it represents the healthy stage. By getting normalized, the remaining data is labeled according to its degradation state. During the healthy stage (pre-FPT), all samples are given the label 1, denoting normal operation. After FPT, the degradation period starts, during which the labels linearly decrease from 1 to 0 as the bearing progresses towards failure.

\subsection{Model Architecture}
The proposed MCSFormer model integrates convolutional processing, and a Swin Transformer backbone in order to model the fault progression dynamics effectively:

\subsubsection{Convolutional Layers}
Horizontal (\(x_{\text{hor}}\)) and vertical (\(x_{\text{ver}}\)) WPD images are processed through convolutional layers consisting of a \(3 \times 3\) convolution with 32 channels, ReLU activation, and \(2 \times 2\) max pooling.

\subsubsection{Feature Fusion and Linear Embedding}
The outputs from the convolutional layers are concatenated to create a unified feature representation. Linear Embedding maps input patches to the embedding space before Swin Transformer processing.

\subsubsection{Swin Transformer Backbone}
The unified features are passed to a Swin Transformer with an embedding dimension of 768 and attention heads \([3, 6, 12, 24]\). The transformer employs shifted window attention to capture local and global dependencies, producing a pooled feature vector.

\subsubsection{Multi-Scale Regression Head}
The pooled features are then fed into a regression head comprising two fully connected layers with ReLU and dropout (\(p=0.3\)), followed by a final layer for RUL prediction.

\subsection{Training and Evaluation}
The model is trained using the Adam optimizer with a learning rate of \(1 \times 10^{-4}\) and a batch size of 16 for 100 epochs.

\subsubsection{Custom Loss Function}
In order to minimize late RUL predictions while generally trying to keep the predicted RUL as close to the actual RUL as possible, a customized loss function introduces an additional penalty term to the normal Mean Squared Error (MSE) loss function for overestimations of RUL while training:
\begin{equation}
\mathcal{L} = \frac{1}{N} \sum_{i=1}^{N} \left( \hat{y}_i - y_i \right)^2 + \lambda \cdot \max(0, \hat{y}_i - y_i)
\end{equation}
where \(N\) is the total number of samples, \(\hat{y}_i\) is the predicted RUL for sample \(i\), and \(y_i\) is the actual RUL. \(\lambda\) controls the penalty magnitude, ensuring the model prioritizes early and accurate predictions.

\subsubsection{Evaluation Metrics}

\paragraph{Mean Absolute Error (MAE)}
 MAE measures the average absolute difference between predicted and actual RUL values.

\paragraph{Scoring Metric}
Based on the scoring metric introduced in the PRONOSTIA dataset manual \cite{nectoux2012pronostia}, the metric used in this work has been modified to align with the labeling strategy. The adapted scoring metric penalizes late predictions more significantly than early predictions due to their higher operational risks. The overall score is calculated as:

\begin{equation}
    \text{Score} = \sum_{i=1}^{N} \left[
    \begin{cases}
        e^{-\frac{\text{Error}_i}{15}} - 1, & \text{if Error}_i < 0 \\[5pt]
        e^{\frac{\text{Error}_i}{5}} - 1, & \text{if Error}_i \geq 0
    \end{cases}
    \right]
\end{equation}

where \( \text{Error}_i = \hat{y}_i - y_i \) represents the prediction error for sample \(i\).

\section{Experimental Results and Analysis}

In this section, the performance of the proposed framework is evaluated through three experiments designed to validate its reliability, generalizability, and improvement in managing late predictions.

\subsection{Experiment 1: Within Operating Condition Testing}
This experiment assesses the model's performance within each operating condition by training on bearings from the same condition while excluding one bearing for testing. The performance is compared with a CNN-SRU model \cite{yao2021}, a Multi-Domain Adversarial Network (MDAN) \cite{zhang2023}, and a Deep Adaptive Transformer model \cite{su2021}.

Table~\ref{tab:within_condition_results} presents the MAE for the competing models. The reported results represent the average performance across all bearings within the same operating condition, when selected as the test bearing.

\begin{table*}[tb]
    \centering
    \renewcommand{\arraystretch}{1.2}
    \setlength{\tabcolsep}{6pt} 
    \caption{Performance Comparison Within Operating Conditions (Average MAE ± Std)}
    \label{tab:within_condition_results}
    \resizebox{0.59\linewidth}{!}{
    \begin{tabular}{l c c c c} 
        \hline
        \textbf{Condition} & \textbf{CNN-SRU} & \textbf{MDAN} & \textbf{Adaptive Transformer} & \textbf{MCSFormer} \\ \hline
        Condition 1 & 0.1797 ± 0.0531 & 0.1548 ± 0.0243 & 0.1175 ± 0.0195 & \textbf{0.0557 ± 0.0194} \\ 
        Condition 2 & 0.2047 ± 0.0783 & 0.1771 ± 0.0663 & 0.0625 ± 0.0316 & \textbf{0.0405 ± 0.0070} \\ 
        Condition 3 & 0.2284 ± 0.0769 & 0.1919 ± 0.0013 & 0.1396 ± 0.0448 & \textbf{0.0896 ± 0.0149} \\ \hline
        \textbf{Average} & 0.2043 ± 0.0694 & 0.1746 ± 0.0306 & 0.1065 ± 0.0320 & \textbf{0.0619 ± 0.0138} \\ \hline
    \end{tabular}
    }
\end{table*}

MCSFormer demonstrates superior performance across all operating conditions with the lowest average MAE. In Condition 1, it performs approximately 69\%, 64\%, and 52\% better than CNN-SRU, MDAN, and the Adaptive Transformer, respectively. In Condition 2, MCSFormer is 35\% better than the Adaptive Transformer, whereas in Condition 3, it is 53\% and 36\% better than MDAN and the Adaptive Transformer, respectively. The overall average MAE also demonstrates its effectiveness, with a 69\% reduction compared to CNN-SRU, 64\% compared to MDAN, and 41\% compared to the Adaptive Transformer. These improvements demonstrate MCSFormer's ability to learn degradation patterns more effectively. While the competing models effectively model the degradation, they lack the same level of robustness and multi-scale feature integration, demonstrating MCSFormer's superior reliability under intra-condition variability.

\subsection{Experiment 2: Cross-Condition Generalization}

This experiment evaluates the generalization capability of MCSFormer by training on 16 bearings across all operating conditions and testing on the remaining bearing. This setup mirrors real-world scenarios where diverse operating conditions contribute to a model’s robustness. The results are compared with Vision Transformer (ViT) and Swin Transformer, both of which have been modified and optimized for this specific task, incorporating the same preprocessing steps used for MCSFormer.

\begin{figure}[b]
    \centering
    \resizebox{0.75\linewidth}{!}{\includegraphics{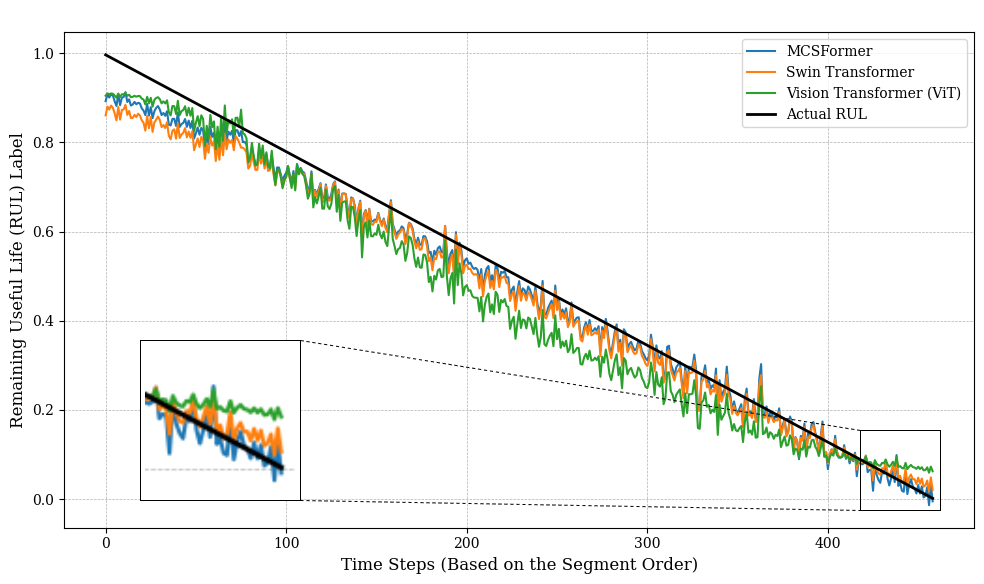}}
    \caption{Comparison of Predicted and True RUL for Bearing1\_1}
    \label{fig:cross_condition_bearing1_1}
\end{figure}

The dataset was split such that 16 bearings were used for training, leaving one bearing for testing. This process was repeated for all 17 bearings (1\_1 to 1\_7, 2\_1 to 2\_7, and 3\_1 to 3\_3). The evaluation metrics include MAE and the scoring metric. A summary of the results for all bearings is presented in Table~\ref{tab:cross_condition_results}. Additionally, the predicted and true RUL curves for Bearing1\_1 are shown in Figure~\ref{fig:cross_condition_bearing1_1}.

\begin{table}[tb]
    \centering
    \caption{Performance Comparison Across All Bearings (MAE and Score)}
    \label{tab:cross_condition_results} 
    \resizebox{0.9\linewidth}{!}{
\begin{tabular}{lllllll}
\hline
\multirow{2}{*}{\textbf{Bearing}} & \multicolumn{2}{c}{\textbf{ViT}}  & \multicolumn{2}{c}{\textbf{SwinT}} & \multicolumn{2}{c}{\textbf{MCSFormer}} \\ \cline{2-7} 
                                  & \textbf{MAE}    & \textbf{Score}  & \textbf{MAE}    & \textbf{Score}  & \textbf{MAE}       & \textbf{Score}    \\ \hline
1\_1                              & 0.0616          & 1.1843          & 0.0436          & 1.1667          & \textbf{0.0351}    & \textbf{1.2130}   \\
1\_2                              & 0.0508          & 0.3439          & \textbf{0.0353} & 0.4362          & 0.0369             & \textbf{0.4585}   \\
1\_3                              & 0.0596          & \textbf{0.8231} & 0.0408          & 0.5629          & \textbf{0.0361}    & 0.6395            \\
1\_4                              & 0.0563          & 0.3155          & 0.0489          & 0.3434          & \textbf{0.0373}    & \textbf{0.3544}   \\
1\_5                              & 0.0542          & 0.6166          & 0.0381          & 0.5137          & \textbf{0.0360}    & \textbf{0.6722}   \\
1\_6                              & 0.0537          & 1.0028          & 0.0459          & 1.0352          & \textbf{0.0362}    & \textbf{1.0863}   \\
1\_7                              & 0.0450          & 0.5481          & 0.0587          & \textbf{0.9254} & \textbf{0.0390}    & 0.6183            \\ \hline
2\_1                              & 0.0427          & 0.1555          & \textbf{0.0294} & \textbf{0.1693} & 0.0336             & 0.1647            \\
2\_2                              & \textbf{0.0297} & 0.2194          & 0.0466          & 0.2237          & 0.0450             & \textbf{0.2542}   \\
2\_3                              & 0.0390          & 0.8258          & 0.0381          & 0.8651          & \textbf{0.0290}    & \textbf{0.9324}   \\
2\_4                              & 0.0653          & 0.3272          & 0.0678          & 0.3441          & \textbf{0.0567}    & \textbf{0.3642}   \\
2\_5                              & 0.0538          & 0.7984          & 0.0442          & 0.5692          & \textbf{0.0377}    & \textbf{0.8677}   \\
2\_6                              & 0.0472          & 0.0445          & 0.0522          & 0.0432          & \textbf{0.0237}    & \textbf{0.0541}   \\
2\_7                              & 0.0539          & 0.0317          & 0.0671          & \textbf{0.0550} & \textbf{0.0322}    & 0.0331            \\ \hline
3\_1                              & \textbf{0.0339} & 0.1582          & 0.0342          & 0.1877          & 0.0358             & \textbf{0.2065}   \\
3\_2                              & 0.0736          & 1.1806          & 0.0641          & \textbf{1.3289} & \textbf{0.0607}    & 1.2944            \\
3\_3                              & 0.0985          & 0.0705          & 0.0931          & 0.0734          & \textbf{0.0915}    & \textbf{0.0969}   \\ \hline
\textbf{Average}                  & 0.0540          & 0.5086          & 0.0499          & 0.5202          & \textbf{0.0413}    & \textbf{0.5476}   \\ \hline
\end{tabular}
}
\end{table}

The results in Table~\ref{tab:cross_condition_results} show that MCSFormer outperforms ViT and Swin Transformer in the majority of bearings (9 out of 17) with reduced MAE and increased score. The superior performance highlights the effectiveness of hierarchical attention mechanisms and feature fusion in handling diverse operational conditions.

Figure~\ref{fig:cross_condition_bearing1_1} further illustrates the predictive performance for Bearing1\_1. MCSFormer's RUL prediction is extremely close to the actual RUL trajectory, reducing late predictions, while ViT and Swin transformer have more noticeable deviations, specifically, towards the end-of-life period which is magnified in the figure for better visualization. Although both ViT and Swin Transformer utilize the preprocessing pipeline effectively, MCSFormer presents the best overall performance with better degradation dynamics modeling for various bearings.

\subsection{Experiment 3: Impact of Custom Loss Function}

To investigate the impact of the custom loss function, the experiment was conducted under conditions similar to the second experiment, where the model was trained on 16 bearings and tested on the remaining one. The model’s performance with the custom loss function was compared to the performance when using the MSE loss function as a baseline. 

While the custom loss function slightly increases the MAE it significantly improves the scoring metric, reflecting a reduction in late predictions while keeping the prediction accurate. This improvement is particularly crucial for real-world applications where safety is a priority. As an example, for Bearing2\_3, by using the custom loss function, the MAE is increased from 0.0258 to 0.0290, and scoring metric improved from 0.8775 to 0.9324.

\begin{figure}[b]
    \centering
    \includegraphics[width=0.75\linewidth]{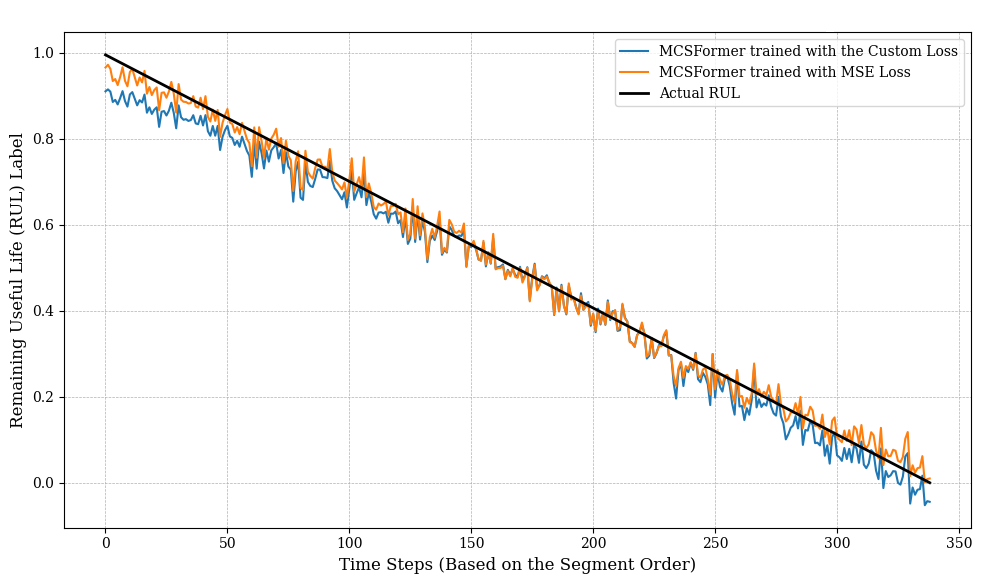}
    \caption{Performance With the Custom Loss vs MSE Loss (Bearing2\_3)}
    \label{fig:custom_loss_comparison}
\end{figure}

Figure~\ref{fig:custom_loss_comparison} provides a visual comparison of the predicted RUL curves with the custom loss function and with the MSE loss function. The model trained with the custom loss demonstrates optimal alignment with the actual RUL trajectory, especially in the later stages of degradation, ensuring fewer late predictions. While the slight trade-off in MAE suggests a marginal reduction in raw accuracy, the improved score highlights the model’s enhanced ability to differentiate between early and late predictions, prioritizing safety over minor changes in numerical precision.

\section{Conclusions and Future Directions}

This study proposes MCSFormer, an optimized multi-channel Swin Transformer model for rolling bearing RUL prediction, utilizing wavelet-based denoising, WPD, and a Swin Transformer backbone. In three comprehensive experiments, MCSFormer demonstrated robust feature extraction and high generalizability under different operating conditions. In the first experiment, its performance in within-condition testing outscored CNN-SRU, MDAN, and Adaptive Transformer in all cases with an average of 69\%, 64\%, and 41\% improvement in terms of MAE, respectively. In the second experiment, its cross-condition generalization was evaluated, where MCSFormer outperformed ViT and  Swin Transformer, modified for this task, both in terms of MAE and scoring metric for 9 out of 17 bearings in the dataset, demonstrating its adaptability to diverse fault progression patterns. 

The third experiment investigated the impact of a custom loss function designed specifically to penalize late predictions more heavily than early predictions. While the custom loss resulted in a slightly larger MAE, it notably decreased the number of late predictions, which is essential for guaranteeing timely maintenance actions. This demonstrates MCSFormer's capacity for balancing predictive precision and operational safety, rendering it a reliable solution for practical predictive maintenance. Future work can explore transfer learning strategies to enhance model robustness when testing on unseen bearings with disparate distributions of data. Analysis of different segmentation techniques, such as sliding and expanding window techniques, could also provide insights into optimizing temporal dependencies for degradation modeling improvement.

\section*{Acknowledgment}

We would like to express our gratitude for the financial support provided
by the Natural Sciences and Engineering Research Council of Canada (NSERC), [NSERC Discovery Grant No. RGPIN-2023-05408].


\end{document}